\documentclass{elsarticle}
\journal{ACM Transactions on Internet Technology}
\usepackage{geometry}
\usepackage{bm}
\usepackage{setspace}
\usepackage{graphicx}
\usepackage{float}
\usepackage{subfigure}
\usepackage{booktabs}
\usepackage{array}
\usepackage{amsthm}
\usepackage{amsmath}
\usepackage{amssymb}
\usepackage{threeparttable}
\usepackage{algorithm}
\usepackage{algorithmic}
\usepackage{subfigure}
\usepackage{multirow}
\usepackage{slashbox}

\theoremstyle{definition}

\graphicspath{{Images/}}
\begin{document}
\begin{spacing}{1.15}
\begin{frontmatter}



\title{\textbf{Exploration and Improvement of Nerf-based 3D Scene Editing Techniques}}
\author[ads]{Shun Fang}\ead{fangshun@acm.org}
\author[ads]{Xing Feng\corref{cor1}}\ead{fengxing@lumverse.com}
\author[ads]{Ming Cui}
\author[ads]{YaNan Zhang}
\cortext[cor1]{Corresponding author: Xing Feng}
\address[ads]{Shijingshan District,\\Beijing Lumverse Technology Co., Ltd, Lumverse Reserch Institute, Bei Jing, 100043, China}

\begin{abstract}
NeRF's high-quality scene synthesis capability was quickly accepted by scholars in the years after it was proposed, and significant progress has been made in 3D scene representation and synthesis. However, the high computational cost limits intuitive and efficient editing of scenes, making NeRF's development in the scene editing field facing many challenges. This paper reviews the preliminary explorations of scholars on NeRF in the scene or object editing field in recent years, mainly changing the shape and texture of scenes or objects in new synthesized scenes; through the combination of residual models such as GaN and Transformer with NeRF, the generalization ability of NeRF scene editing has been further expanded, including realizing real-time new perspective editing feedback, multimodal editing of text synthesized 3D scenes, 4D synthesis performance, and in-depth exploration in light and shadow editing, initially achieving optimization of indirect touch editing and detail representation in complex scenes. Currently, most NeRF editing methods focus on the touch points and materials of indirect points, but when dealing with more complex or larger 3D scenes, it is difficult to balance accuracy, breadth, efficiency, and quality. Overcoming these challenges may become the direction of future NeRF 3D scene editing technology.
\end{abstract}

\end{frontmatter}

\section{Introduction}
NeRF (Neural Radiance Fields)\cite{mildenhall2021nerf} was first conceptualized in the best paper at ECCV 2020, proving the ability to build high-quality 3D results popularized by generating photorealistic images rendered from new viewpoints for a given sparse set of images. In recent years, NeRF has become a popular method for new perspective synthesis. Many scholars have explored, researched, and improved the technique to become one of the fundamental paradigms in the field of 3D vision in only two years. NeRF quickly adapted to a wider range of application areas, including tasks such as 3D scene reconstruction, rendering, localization,and generation.

Efficiently and accurately decomposing the geometric structure of a 3D scene and arbitrarily editing it is a key issue for 3D scene understanding and interaction, and the basis for applications such as virtual reality and intelligent machines. With the increasing popularity of efficient 3D reconstruction techniques such as Neural Radiation Field (NeRF), it is becoming increasingly feasible to generate realistic synthetic views of real-world 3D scenes. At the same time, the demand for 3D scene manipulation is rapidly increasing due to the wide range of content re-creation use cases. Many recent studies have introduced extensions for editing NeRFs, such as manipulating explicit 3D representations, training models to enable color modification or removal of certain objects, or text-to-3D synthesis using diffusion models. These advances extend the versatility of implicit volumetric representations.

In the field of 3D scene editing, classical geometric methods such as SfM/SLAM can only reconstruct sparse and discrete point clouds, which cannot contain geometric details and cannot understand and manipulate the scene. NeRF can realize clearer and smoother effects for 3D scene editing based on its micro-able advantage. Examples include manipulating explicit 3D representations, training models for color modification or removal of certain objects, or text-to-3D synthesis using diffusion models. These advances extend the versatility of implicit volumetric representations. However, although the current NeRF-based editing of 3D scenes has been more richly explored, such as changing the colors and shapes of scenes or objects, intuitively and efficiently editing NeRF scenes is still an outstanding challenge due to the high computational cost requirements of NeRF, which also leads to the need for complex and computationally expensive training procedures for each editing task. In order to sort out the recent explorations of NeRF in the field of 3D scene editing and the urgent challenges to be solved, the initial exploration of scene and object editing based on NeRF, the generalization of more capabilities by combining with the existing mature models, and the reconstruction and editing of light and shadow will be explored in three aspects.

\section{Initial exploration of NeRF scene/entity editing}

Traditional Neural Radiation Field fitting or generation of a scene or object, rendering a new view of a particular scene, but not editing it. Initial exploration in the direction of NeRF editing of scenes/objects using different networks and implicit vectors to decouple the shape and appearance; user editing on 2D rendered images, backpropagation optimization or forward editing using the networks and implicit vectors to further improve the final visual quality.

EditNeRF implements an implicit continuous volumetric representation of 3D objects that can be edited and controlled by the user, allowing scene editing using conditional reflections from the user's input image, whereby the user can change the color of a localized appearance, perform local shape removal on a 3D object, or transfer colors and shapes from a target object instance. Edit-NeRF's shape representation consists of category-specific shared shapes network $F_{share}$ and an instance-specific shape network $F_{inst}$ that are composed of Groups are defined in terms of $z_s$ conditioned on the category, whereas $F_{shared}$ is not. Theoretically, the $F_{shared}$ behaves as a deformation field, the NeRF editing problem is defined using GLO as the NeRF network parameters and potential codes $z_s$ that $z_a$ The joint optimization problem of the NeRF luminosity loss optimization is then performed on the latent codes, followed by optimization of the MLP weights, and finally optimization of the latent codes and weights seeks to enable editing of the entire shape class (shape class)\cite{liu2021editing}, NeRF-Editing subsequently implements a user-editable rigid transformation model of the object and this effect can be applied to arbitrary objects\cite{yuan2022nerf}, NeuMesh on This foundation oriented to geometry and texture editing of de-entangled neural mesh implicit field learning, which enables NeRF models to be more adaptable to fine-grained editing of general objects in daily life, including geometric editing, specified texture editing with texture swapping, filling and painting operations, etc.\cite{yang2022neumesh}, which further improves the quality of the representation and editing capabilities; and ARF implements scene editing in the stylized dimension\cite{zhang2022arf}. object -NeRF can separate objects from backgrounds, and designs a new two-path architecture to learn the object combination neural radiation field given a collection of posed images and a 2D instance mask, to support high-quality new view rendering and object manipulation, and to enable editing of real scenes.

\section{Generalizable NeRF editing}
The single, plain NeRF model requires intensive shooting of each object or scene as well as independent training, while many current studies have extended the editing performance of NeRF with the advantages of GaN, and Diffusion model, in face editing and multimodal generation.

\subsection{NeRF extends 3D-aware GANs portrait editing capabilities}
Stylegan-based methods have better results for synthesizing clear 2D face portraits by learning to edit attribute-specific orientation potential spaces, or by learning more unraveled and controllable potential spaces various priori, but direct application to editing different views of a 3D face presents the problem of trying to be inconsistent.

Currently 3D-aware GANs constructed using implicit representations have gained more attention, and methods such as PiGAN\cite{chan2021pi}, a GaN-based 3D face editing work, have been able to generate rich 3D faces, but some of these methods are memory inefficient and unable to perform fine-grained editing of the model. FENeRF\cite{sun2022fenerf} attempted to enable local face editing in and the results showed insufficient photorealism. In addition, optimization-based GAN inversion is very time-consuming, and facing real-time interactive editing tasks still has difficulties that need to be overcome. Recent research has mainly used a hybrid approach of neural style migration + generative adversarial network (GAN) to allow for clearer and more realistic editing of 3D facial images with richer functional dimensions that can be generalized. Some of the representative results include:

IDE3D\cite{sun2022ide} proposes a generative neural semantic field with decoupled geometry and material to align 3D semantics and geometry by additionally outputting a semantic mask in a ge  ometric branching network; the editing principle is that 2D semantic graph editing is mapped to the semantic field, thus editing the 3D semantics and the geometry aligned with it. The end result supports interactive 3D face compositing and localized editing, specifically supporting a variety of efficient free-view portrait editing tasks.

AnifaceGAN\cite{wu2022anifacegan} proposes an animatable 3D-aware GAN for multi-view consistent face animation generation. The key idea is to decompose the 3D representation of 3D-aware GANs into a template field and a deformation field, where the former represents different identities with typical expressions and the latter characterizes the expression changes of each identity. In order to achieve meaningful control over facial expressions through deformation, the method proposes a 3D-level imitation learning scheme between the generator and the parameterized 3D facial model during adversarial training of 3D-aware GANs. This helps the method to achieve high-quality animatable face image generation with strong visual 3D consistency, even when trained only with unstructured 2D images.

Next3D\cite{sun2023next3d}:A dynamic 3D planar representation based on neural texture mapping is proposed, which divides the whole head into dynamic and static parts and models them separately. For the dynamic part, a new representation combining fine-grained expression control for mesh-guided explicit deformations and implicitly proposes a generative texture rasterized tri-plane, which learns facial deformations from generative neural textures at the top of the parametric template mesh and samples them into three orthogonal views and axis-aligned feature planes by standard rasterization, forming a tri-plane feature representation including expression, blink, gaze direction and full head pose.

From GAN-based methods to neural style migration and hybrid models, the current results innovate many new models and methods that make efforts in the intuitiveness of editing 3D semantics and geometry, the accuracy of editing control of 3D models in special domains, and the ability to further edit the dynamic and static features of 3D models and have been applied mainly in the field of face editing.

\subsection{Diffusion-based multimodal exploration of NeRFs}
Diffusion model is a class of generative methods that has received much attention in recent years. The forward diffusion process adds Gaussian noise to some input image/feature mapping in some T-steps. The reverse process is generative and can be used to create images from Gaussian noise. The diffusion model is used to continuously iterate the editing training set while optimizing the neural radiation field parameters so that the NeRF rendering results converge with the edited images generated from the given text. The diffusion model provides a high degree of control over generation by using domain-specific encoders that allow the use of both text and image-based cues. Recent research has produced diffusion NeRF models by training the NeRF model from scratch on images generated by diffusion methods, allowing the model to generate views of the same object in different poses with appropriate textual cues.

Instruct3D-to-3D\cite{kamata2023instruct} proposes a high-quality 3D-to-3D conversion method that converts a given 3D scene to another one based on textual instructions. The method applies a pre-trained image-to-image diffusion model for 3D-to-3D conversion. This makes it possible to maximize the likelihood of each viewpoint image and high-quality 3D generation. Furthermore, the method explicitly takes the source 3D scene as a conditional input, which enhances 3D consistency and controllability of the source 3D scene structure. The authors also propose dynamic scaling, which allows to adjust the intensity of the geometry transformation. The authors perform quantitative and qualitative evaluations, which show that the proposed method achieves higher quality 3D-to-3D transformations than the baseline method.

InstructNeRF2NeRF\cite{haque2023instruct} proposes a pairwise alignment technique that extracts fields from an already sought NeRF model (which MEASURES the likelihood of a point being on the surface of an object). Then using nerf2nerf registration as a robust optimization, iteratively find a rigid transfromation to align the surface fields of the two scenes.

Edit-DiffNeRF\cite{yu2023edit} is based on the diffusion model of the text-generated graph, and utilizes text to achieve more intuitive and interactive 3D or 4D editing of NeRF. The diffusion model is used to iteratively edit the training set and optimize the neural radiation field parameters, so that the NeRF rendering results and the editing image generated from the given text tend to be consistent. The existing diffusion model can only edit and generate 2D images, and with the help of the dynamic NeRF, it can be upgraded from 2D to 4D, and achieve high-quality and consistent 4D editing generation.

\subsection{Exploration of 4D generation by combining multiple models}
Existing diffusion models can only edit and generate 2D images, but with dynamic NeRF, editable images can be upscaled from 2D to 4D, which in turn can realize high-quality 4D generation. Currently, studies have been conducted to explore the combination of multiple models to realize 4D generation and editing by NeRF.

Control4D\cite{shao2023control4d}: Combine Tensor4D with GAN to realize a 4D GAN, using 4D GAN to learn the image distribution generated by diffusion model in different moment viewpoints, avoiding the direct supervision of the image so as to achieve high quality editing and generating effect, the supervised signal generated by 4DGAN discriminator is smoother compared with that of the diffusion model, which makes the temporal and spatial editing of 4D scene more consistent and the network converges more efficiently. The 4DGAN discriminator produces smoother supervised signals than the diffusion model, resulting in better spatio-temporal consistency of 4D scene editing and faster network convergence.

\section{Light and shadow editing}
While expanding the modalities and dimensions of NeRF, many studies are also expanding the material information of NeRF representations, and exploring the practical dimensions of NeRF from the perspective of light and shadow editing. The initial research\cite{zhang2021nerfactor,zhang2021physg,zhang2022modeling}, such as the idea is to decompose NeRF color representation into Geometry (Normal) + Material (BRDF) + Lighting, and re-combine the rendering to achieve heavy lighting and material editing. At present, the editing of NeRF light and shadow information mainly focuses on indirect light simulation and optimization of scene representation details.

\subsection{Indirect light simulation}
Although the initial NeRF light and shadow editing studies have been able to isolate the light for individual editing, it is difficult to explain the complex light transport effects by simulating only direct light. The current study attempts to use NeRF irradiance modes in different directions to indirectly illuminate the light, and to create more detailed light modes by simulating light bouncing multiple times, and to achieve visual quality light editing based on physical rendering.

TensorIR\cite{jin2023tensoir}:proposes a tensor decomposition-based inverse rendering method that simultaneously implements physical model-based estimation and radiation field reconstruction, leading to realistic novel view synthesis and relighting results. Unlike the previous use of purely MLP-based NeRF, on top of TensoRF, for radiation field modeling, the estimated scene geometry, surface reflectance and ambient illumination are derived from the conditions of multi-view images taken under unknown illumination. Benefiting from TensoRF, the method can efficiently represent auxiliary coloring effects (e.g., shadows and indirect lighting) in images captured under single or multiple unknown illumination conditions. The low-rank tensor representation allows the method not only to achieve fast and compact reconstruction, but also to capture lighting conditions better using any number of shared information.

NeFII\cite{wu2023nefii}: Mainly to address the shortcomings of current methods in handling high-frequency reflections, an end-to-end inverse rendering pipeline is proposed, which combines Monte Carlo sampling with a spherical Gaussian function to realize high-quality rendering. The method estimates geometry, material and illumination in multi-view RGB images via an inverse rendering ground-truth approach while considering near-field indirect lighting, while introducing Monte Carlo sampling-based path tracking and caching indirect lighting as neural luminance to achieve a physically faithful and easily optimized inverse rendering method. In addition, for efficiency and practicality, a Gaussian distribution is utilized to represent smooth ambient illumination, and ultimately higher rendering results are achieved through path-tracing results of unobserved radiation, and joint optimization of materials and illumination.

\subsection{Detail-oriented scene representation}
Traditional surface-based rendering methods struggle to simulate complex geometries and transparent materials. Recovering the physical properties of an object's appearance from images taken under unknown illumination is challenging, but essential for photorealistic rendering. By combining a microplane model with a neural radiation field to simulate the interaction of light rays with the microplane during transmission, an implicit microplane field is created, and body rendering is used to achieve high-quality rendering that does not depend on geometric surfaces.

NMF\cite{mai2023neural}: proposes the microfacet reflectance material model Neural Microfacet Fields, a method for recovering materials, geometry (bulk density) and ambient lighting from a collection of scene images, using Monte Carlo sampling and ray tracing to simulate mutual reflections. Specifically the method applies a microfacet reflectivity model in a volumetric setup by treating each sample along the ray as a surface rather than an emitter. The use of surface-based Monte Carlo rendering in the volumetric setup allows the method to efficiently perform inverse rendering, which in turn can capture high-fidelity geometry and high-frequency illumination details.

NeMF\cite{zhang2023nemf}: This approach addresses the problem that current work cannot well handle the detailed representation of scenes with very complex geometries, translucent objects, etc. It simulates the reflection and refraction effects of light at microslice by combining the microslice volume model with the neural radiation field method. Specifically, the method performs inverse volume rendering by representing the scene using a micro-sheet volume, which, in contrast to surface-based rendering, assumes that the space is filled with an infinite number of small sheets and that light is reflected or scattered at each spatial location according to the micro-sheet distribution. The method further employs a coordinate network to implicitly encode the volume of the microflakes and develops a differentiable microflake volume renderer that in principle trains the network in an end-to-end manner. The method focuses on recovering the appearance properties of more complex geometries and scattering objects, and realizes relighting of shuttering details, and material editing.

\section{Summary and outlook}
During the development of NeRF, initial explorations focused on editing scenes and objects. Using the microscopic nature of NeRF, researchers have explored methods for editing color, shape, and texture, making it possible to perform more intuitive and accurate editing of 3D scenes. These explorations include editing using conditional radiation fields and implicit volume representations, enabling object rigidity transformations, decoupled editing of geometry and texture, and finer geometric and texture manipulations. Meanwhile NeRFde generalizable editing capabilities have also received attention. Researchers have worked on realizing the broader applicability of NeRF, trying to extend the editing domain to areas such as 3D-to-3D transformation and animation generation. From GAN-based methods to hybrid models, new models and methods are constantly innovated, which make editing 3D semantics and geometry more intuitive and precise, and are widely used in areas such as face editing. In addition, Diffusion-based NeRF multimodal exploration and 4D generation by combining multiple models are also hot spots in current research. These methods achieve higher quality and more consistent edit generation by extending the editing dimensions of NeRF. Through methods such as Diffusion model, researchers have realized edit generation from 2D to 4D, which further enriches the application scenarios of NeRF. In the development of NeRF, light and shadow editing has also become an important research direction. By decomposing the color representation of NeRF, researchers try to reconstruct the lighting information of the scene and achieve more detailed editing of light and shadow. These methods include indirect light simulation and detail-oriented scene characterization, which are dedicated to simulating the complex transmission effects of light and achieving high-quality rendering.

However, most of the current NeRF-based editing methods only focus on the lighting and material composition of the scene, facing more complex 3D scenes or larger scene editing and reconstruction, still can not take into account the accuracy and breadth, efficiency and quality, one by one to overcome the above challenges may be beneficial to become the future direction of exploration of the NeRF 3D scene editing technology. At the same time, the recent innovative method of combining generative models is also gradually expanding the diversity of NeRF rendering scenes, and NeRF technology is expected to be further developed, especially in the multi-modal, optimized model accuracy of the controllable editing and the expansion of application areas. As the research on NeRF continues to deepen, it will be more widely used in virtual reality, intelligent machines and other fields, bringing more possibilities for 3D scene understanding and interaction.

\section{Acknowledgments}
We thank the Beijing Municipal Science and Technology Commission and the Zhongguancun Science and Technology Park Management Committee for their support under project numbers Z221100007722005, 20230468276.

\bibliographystyle{elsarticle-num}
\bibliography{mybib}
\end{spacing}
\end{document}